\documentclass[conference]{IEEEtran}
\IEEEoverridecommandlockouts
\usepackage{cite}
\usepackage{lipsum}

\usepackage{amsmath,amssymb,amsfonts}
\usepackage{algorithmic}
\usepackage{graphicx}
\usepackage{textcomp}
\usepackage{xcolor}
\usepackage{xcolor,colortbl}
\usepackage{threeparttable}

\definecolor{Gray}{gray}{0.93}
\definecolor{LightCyan}{rgb}{0.88,1,1}

\newcolumntype{a}{>{\columncolor{Gray}}c}
\newcolumntype{b}{>{\columncolor{white}}c}

\begin{document}

\title{Domain Adaptive Transfer Learning for Fault Diagnosis\\
{\footnotesize }
\thanks{This research was funded by the Swiss National Science Foundation (SNSF) Grant no. PP00P2\_176878.
}
}

\author{\IEEEauthorblockN{Qin Wang}
\IEEEauthorblockA{
\textit{ETH Z\"{u}rich }\\
Z\"{u}rich, Switzerland \\
qwang@ibi.baug.ethz.ch}
\and
\IEEEauthorblockN{Gabriel Michau}
\IEEEauthorblockA{
\textit{ETH Z\"{u}rich }\\
Z\"{u}rich, Switzerland \\
michau@ibi.baug.ethz.ch}
\and
\IEEEauthorblockN{Olga Fink}
\IEEEauthorblockA{
\textit{ETH Z\"{u}rich }\\
Z\"{u}rich, Switzerland \\
fink@ibi.baug.ethz.ch}
}

\maketitle

\begin{abstract}
Thanks to digitization of industrial assets in fleets, the ambitious goal of transferring fault diagnosis models from one machine to the other has raised great interest. Solving these domain adaptive transfer learning tasks has the potential to save large efforts on manually labeling data and modifying models for new machines in the same fleet. Although data-driven methods have shown great potential in fault diagnosis applications, their ability to generalize on new machines and new working conditions are limited because of their tendency to overfit to the training set in reality. One promising solution to this problem is to use domain adaptation techniques. It aims to improve model performance on the target new machine. Inspired by its successful implementation in computer vision, we introduced Domain-Adversarial Neural Networks (DANN) to our context, along with two other popular methods existing in previous fault diagnosis research. We then carefully justify the applicability of these methods in realistic fault diagnosis settings, and offer a unified experimental protocol for a fair comparison between domain adaptation methods for fault diagnosis problems.  

\end{abstract}


\section{Introduction}

Digitization of industrial assets gives modern maintenance systems access to larger amount of condition monitoring data at a lower cost. With the help of increased availability of these collected data, data-driven fault diagnosis methods have shown great potential in extracting system health information from complex data of varied nature. In recent years, building on the success of data-driven methods, the ambitious goal of transferring fault diagnosis models from one machine to the other has raised great interests. If such a problem is solved, the industry can save a considerable amount of effort on manually labeling data and modifying models for new machines in the same fleet. A successful solution to the problem can potentially save both time and fortune for the industry. One underlying problem of data-driven methods on these transferring tasks, as with many other application areas of data-driven methods, is its strong requirement on the quality of data. The lack of representativeness of the training data can dramatically affect the model performance on the target machine. If the target machine operates on a different working condition other than the one observed in training data, the model performance may degrade dramatically.

Deep neural networks, as one of the popular data-driven methods, especially suffer from this problem. Recent research~\cite{zhang2016understanding} has shown that deep networks are able to memorize the entire data-set even when random labels are given. This strong capacity of memorization can lead to poor generalization performance on new machines as well as on new operating conditions. Deep models are thus likely to overfit to the given training set that might be unrepresentative and cannot generalize well to new data in reality. Specifically, in a fault diagnosis context, a carefully trained deep model is likely to degrade on a newly deployed machine in the same fleet because of different environmental and operating conditions. Another special case is when operating conditions of a machine change over time, deep models are likely to classify new operating conditions as faults simply because it has not observed similar patterns in training period. 

One intuitive solution to this problem is to add data under new operating conditions to the training set, and re-train the model. However, this is usually infeasible due to the fact only limited data are available under new operating conditions and new machines, and the reality that these new data are often unlabeled makes the problem even harder.  

Domain adaptation methods are designed to tackle this kind of dilemma where two different machines are involved. Domain adaptation methods aim to leverage a small amount of unlabeled data under new operating conditions, and improve the model's generalization ability. As the methods aims at transferring results achieved on a first domain with labeled data under given operating conditions, to a second domain with unlabeled data and different operating conditions, it is referred to as "domain adaptation". It has been a widely discussed topic in fields such as computer vision and natural language understanding~\cite{pan2011domain, long2015learning, li2016revisiting, ganin2014unsupervised,  saito2018maximum}. Inspired by the successful implementation of Domain-Adversarial Neural Networks (DANN)~\cite{ganin2014unsupervised}, we propose to make use of its ability to alleviate domain difference for fault diagnosis problems.

\begin{figure}
	\centering
	{\includegraphics[width=\columnwidth]{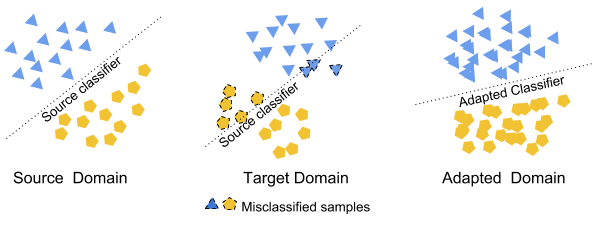}}
	
	\caption{Domain Adaptation in a Toy Example.}
	\label{toy}
\end{figure}
Over the last few years, several fault diagnosis papers~\cite{zhang2017new, zhang2018adversarial, li2019multi} also proposed to apply other domain adaptation methods to improve model performance on new operating conditions. These recent attempts raise a natural question: Are domain adaptation methods applicable in realistic fault diagnosis settings? How well do they perform comparing to each other? In this paper, we argue that previous papers have not answered these questions sufficiently. A fair evaluation across different methods requires careful choice of network structures, data preprocessing, training strategy, etc. The aim of this paper is to answer these questions by using a unified experimental protocol on a popular dataset, the Case Western Reserve  University (CWRU) dataset for rolling element bearings in rotating machinery. We believe the proposed protocol shows the future potential of domain adaptation methods in fault diagnosis. 

%The focus of this work is as follows:
%\begin{itemize}
%	\item 
%\end{itemize}

\section{Related Work}
Deep learning methods~\cite{li2015multimodal, zhai2016deep, jia2016deep, tamilselvan2013failure}, have attracted large amount of attention by promising better performance without the need of hand-craft features. However, it is known that when a trained model is deployed on unseen operating conditions, the performance can deteriorate dramatically because of the operating condition difference, in other words, data distribution difference, between training and testing machines. 

In previous works~\cite{li2018cross, zhang2018deep}, this difference is often called domain shift, where training data under observed operating conditions are considered as source domain, and newly collected data under new operating conditions are considered as target domain. The domain shift problem has been widely discussed in other fields such as computer vision~\cite{saenko2010adapting, tzeng2017adversarial }. To alleviate the effect of domain shift in the input space, one motivation is to align the distributions in intermediate feature space, this intuition leads to a series of domain adaptation methods~\cite{pan2011domain, long2015learning, li2016revisiting, ganin2014unsupervised,  saito2018maximum}. For example, \cite{pan2011domain} proposes to learn transfer components across domains. Deep Adaptation
Network (DAN) method~\cite{long2015learning} proposes to minimize domain discrepancy by minimizing the Maximum Mean Discrepancy (MMD) between source and target layers. Driven by similar motivation,  Adaptive Batch Normalization (AdaBN)~\cite{li2016revisiting} aligns the distributions through a modified batch normalization layer and calculate batch normalization statistics separately for source and target data. Along with the success of adversarial training on other tasks, DANN~\cite{ganin2014unsupervised} proposes to align the distributions by adopting a domain discriminator and training the model adversarially. Recently, \cite{saito2018maximum} proposes to align distributions of source and target by utilizing the task-specific decision boundaries, and maximizing classifier discrepancy. 

For fault diagnosis applications, existing papers usually focus on the case where unlabeled data in target domain are fully provided, and directly apply the above domain adaptation techniques to solve the problem. For example, \cite{zhang2017new} proposes to use AdaBN to learn a model with good anti-noise and domain adaptation ability on raw vibration signals. Similarly, \cite{zhang2018adversarial} propose to align the distributions of intermediate layers between source and feature extractors by adversarial training. \cite{michau2019} consider the problem of fault detection within a fleet using unsupervised feature alignment. Recently, \cite{li2019multi} uses MMD-minimization to align the full source and target distributions for rotationary machines.

%\begin{equation}
%a+b=\gamma\label{eq}
%\end{equation}
%\eqref{eq}
\section{Problem Description}
% Definition, two machines similar characteristics, different working conditions. 
The main motivation behind domain adaptation in fault diagnosis is that, in industry, it is not uncommon to see a fleet of similar machines with similar purposes available. It would be beneficial to manually label the data from one single machine and later transfer model knowledge from this well-studied machine, to other newly deployed machines in the same fleet, given that machines in the same fleet share characteristics and features. However, the fact that these machines may be operated under different conditions, and not even necessarily by a single operator, makes the transfer hard in reality. This change of operating condition, can be described as the distribution difference between training and testing data. 

Besides learning from labeled data from source machine, domain adaptation aims to leverage the limited data from target machine and try to improve the performance on target machine by taking these partial data from target machine into consideration. Under the ideal scenario, this should help the model to perform better on the target machine. 

To evaluate the effectiveness of different domain adaptation methods in fault diagnosis applications, following the setup of most previous papers, we propose the following set up, based on how the fault diagnosis transfer problem with two machines had been formulated in previous papers.The first machine, denoted by source, has been operating for a long time. This made possible the collection of representative data on different faults. The second machine, denoted by target, has less data available, and they are unlabeled. The source and target machines share similar characteristics but are operating under different operating conditions. We further assume that these two machines share the same sets of fault types. The goal of the training is to improve the performance of the model on the target machine. 

\subsection{Domain Adaptation Task}

Formally, we consider our first domain adaptation task for fault diagnosis. Given: 
\begin{itemize}
    \item Labeled training data from source machine $$\mathcal{D}_s=\{(x_{s1}, y_{s1}), ...,(x_{sn}, y_{sn})\},  y_{si} \in Y $$
    \item Unlabeled data from target machine $$\mathcal{D}_{t}=\{x_{t1}, ...,x_{tk}\}$$
\end{itemize}

where $y$ are condition classes to predict, i.e. healthy state and various faulty states, and $Y$ is the union of all possible classes $\{0, ..., C-1\}$. Labels of target data are unavailable during training. The target of the task is to train a model using labeled $\mathcal{D}_l$ and unlabeled $\mathcal{D}_{s}$, and improves its performance on $\mathcal{D}_t$. We denote the ground truth labels as $\{y_{t1}, ...,y_{tk})\}, y_{ti} \in Y$. 

In this setup, we assume that the unlabeled data from target machine already covers most of the fault types, thus the label space is the same between $\mathcal{D}_l$ and $\mathcal{D}_{s}$. This is the setup used by most previous domain adaptation papers in fault diagnosis.

\section{Methods}
We propose to evaluate several popular domain adaptation methods under a unified experimental protocol. In this section, we first introduce the shared backbone architecture we used in all our experiments. Then we introduce the domain adaptation methods that we are going to compare.

\subsection{Baseline Architecture}
One main obstacle on comparing different domain adaptation in fault diagnosis is that different works use different architectures for their experiments, thus direct comparison on results is unfair due to the different capacities of networks. In this paper, we evaluate all domain adaptation methods using the same basic architecture to ensure a fair comparison. 

\begin{figure}
	\centering
	{\includegraphics[width=\columnwidth]{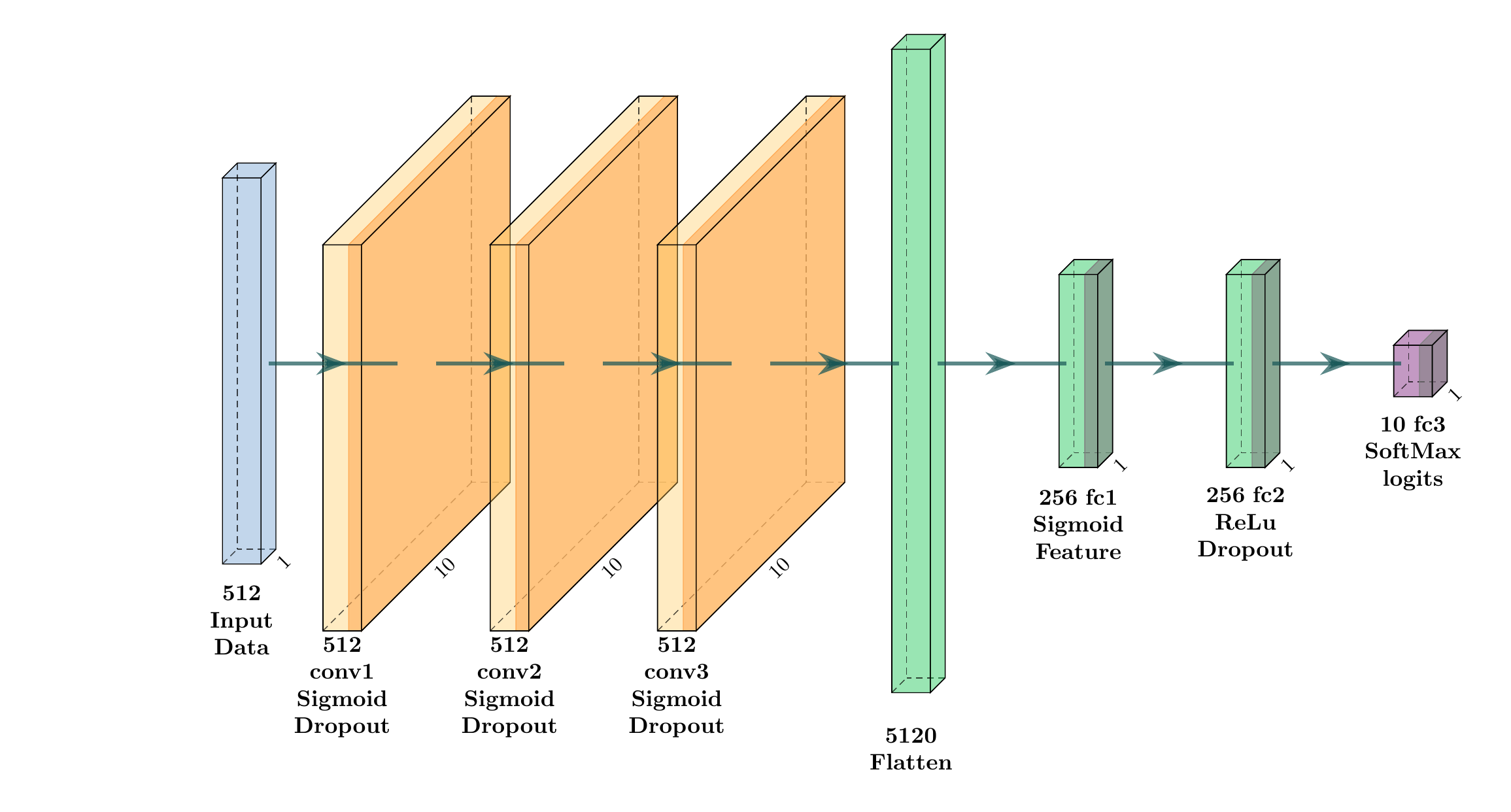}}
	
	\caption{The backbone architecture we used in experiments.}
	\label{feat}
\end{figure}

The CNN backbone from~\cite{li2018cross} is used as shown in Fig~\ref{feat}. The basic architecture is composed of two parts: a feature extractor, and a basic classifier. The feature extractor $f_e$ takes input data and output a feature representation of the given data. It includes three 1-D convolutional layers. Each comes with a filter length of 3, and a hidden size of 10, following the sigmoid activation function, as well as a dropout layer with 0.5 as dropout rate. The representation is then flattened and passed through a fully-connected layer to get mapped into a predefined feature size. Following the original paper, the feature size of 256 is used. 

We choose the architecture in Fig~\ref{feat} because it composes a rather strong baseline for domain adaptation tasks. The effectiveness of the architecture is proved in~\cite{li2018cross}, and also validated by our re-produced results. 

We use a two layer classifier $f_l$ after extracting feature representation of the input data. The first layer is 256 units fully-connected layer with ReLu activation and dropout. The second fully-connected layer then maps the signal into scores for each class. Finally, softmax cross-entropy loss is used for all our experiments. The classification loss shared by all our experiments are thus:
$$\mathcal{L}_{clf} = -\frac{1}{n}\sum_{i=1}^{n}\sum_{c=0}^{C-1} y_{si_c}\log p_{si_c},$$
where $p_{si}$ is the softmax output of the basic backbone.

\subsection{Domain Adaptation Methods}
We now introduce the three domain adaptation methods to be compared. The methods are chosen based on their applicability to deep models. Classic methods such as Transfer Component Analysis (TCA)~\cite{pan2011domain} were not considered because of their inferior performance proved by experiments in~\cite{li2018cross}. To our knowledge, it is the first time DANN method is introduced in a fault diagnosis context.

\subsubsection{Domain-Adversarial Neural Networks (DANN~\cite{ganin2014unsupervised})} 

Since the operating conditions of source and target machines are different, if model is trained naively, it would be easy to distinguish a target machine feature from source. The main idea of adversarial distribution alignment methods is to tackle this problem by making the feature extractor unbiased on features from source and target machines.  This is achieved by an idea closely related to GAN~\cite{goodfellow2014generative}. By adding a discriminator and introducing adversarial training,  DANN~\cite{ganin2014unsupervised} is a method that aligns the source and target feature distributions and makes them hard to be distinguished.  

\begin{figure}
    	\centering
	{\includegraphics[width=\columnwidth]{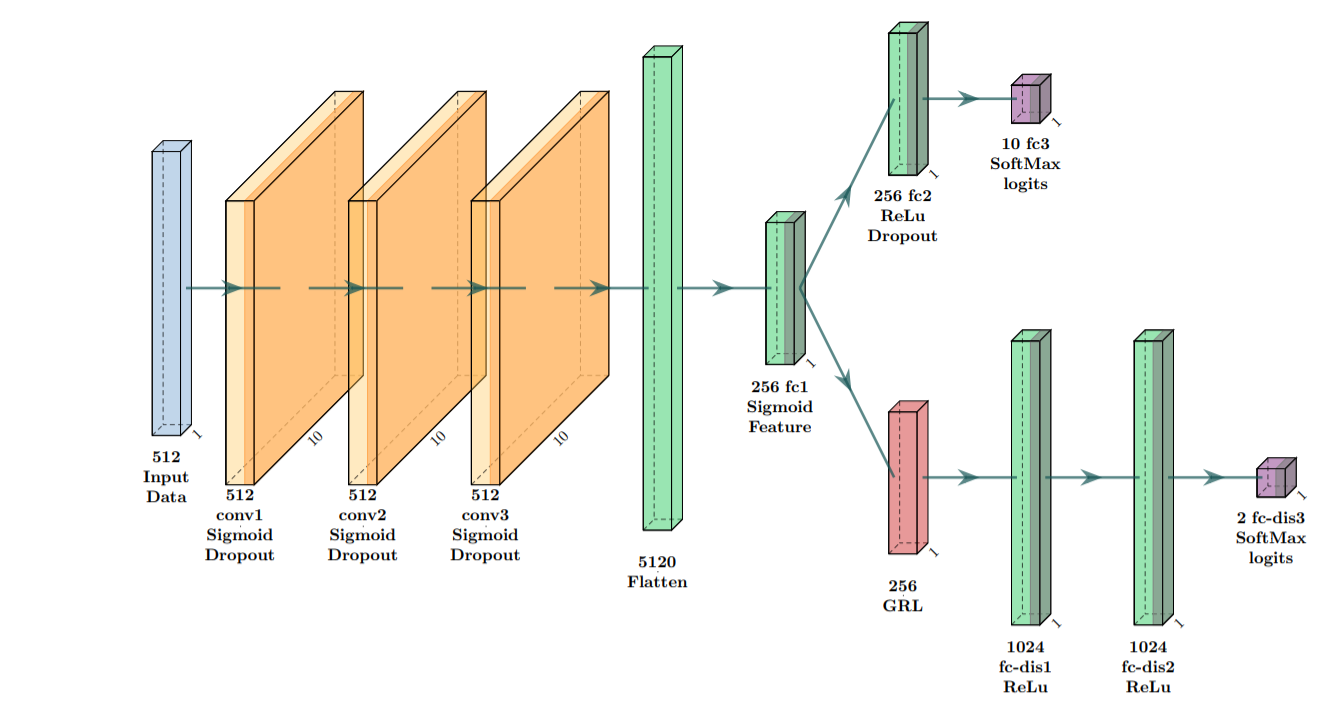}}
	
	\caption{Modified architecture for DANN in comparison experiments.}
	\label{grl}
\end{figure}

Formally, we consider the $\mathcal{H}$-Divergence~\cite{ben2010theory, cortes2011domain} between source and target distributions $\mathbf{L}$ and $\mathbf{UL}$ w.r.t. a hypothesis class $\mathcal{H}$, which is a set of binary classifiers $\eta$:
$$d_{\mathcal{H}}(\mathbf{L}, \mathbf{UL}) = 2 \sup_{\eta \in \mathcal{H}} |P_{x\sim \mathbf{L}}[\eta(x)=1] - P_{x\sim \mathbf{UL}}[\eta(x)=1]|$$

During training, we aim to reduce the $\mathcal{H}$-Divergence between source and target feature distributions. Fortunately, the adversarial alignment method proposed in~\cite{ganin2014unsupervised} for domain adaptation can effectively reducing $\mathcal{H}$-Divergence by reversing gradients and changing the representation space. We modify their method for our semi-supervised learning scenario. 

The neural network includes three component: a feature extractor $f_e$, a label predictor $f_l$, and a discriminator $f_d$.  The divergence reduction is achieved by introducing the discriminator $\theta_d$ to tell whether the features come from source or target data while asking the feature extractor to fool the discriminator. During learning stage, on one side, we are trying to achieve the traditional training objective that minimize the label prediction error. At the same time, we are also pushing the features to be invariant towards its origin, i.e. the divergence between $f_e(x_l)$ and $f_e(x_u)$ to be reduced. This is monitored by the discriminator, where a successful alignment should yield high domain prediction loss. Formally, this is equivalent to the following min-max problem:

%\begin{align}
$$\mathcal{L}(\theta_e,\theta_l,\theta_d) = \mathcal{L}_{clf} (\theta_e, \theta_l) - \lambda_d \mathcal{L}_d(\theta_e, \theta_d)$$
$$(\hat\theta_e, \hat\theta_l) = \arg\min_{\theta_e, \theta_l} \mathcal{L}(\theta_e, \theta_l, \hat\theta_d)$$
$$\hat\theta_d = \arg\max_{\theta_d} \mathcal{L}(\hat\theta_e, \hat\theta_l, \theta_d) $$
%\end{align}

The loss function is divided into two parts, label prediction loss and domain prediction loss. The first term is the usual supervised loss for labeled data, and intends to train the feature extractor and label predictor. The second term is an adversarial loss that ensures the features to be domain-invariant and thus aligns the two distributions. 

In argmin step, we are minimizing the label prediction loss as well as maximizing the domain prediction loss to achieve a domain-invariant features. In maximization step, we are minimizing the domain prediction loss, and thus training the domain predictor to provide precise prediction of the origin of features. This min-max problem is solved by adding a gradient reverse layer between feature layer and discriminator as described in~\cite{ganin2014unsupervised}.

In all our experiments, we use a three layer fully-connected classifier as our discriminator. The first two layers have hidden size of 1024 with ReLu activation, while the last layer maps the signal into 2 classes: source and target. Cross entropy loss is used for the discriminator loss. 

By using gradient reverse layer and the above setup, the loss function can be reformulated into:
$$\mathcal{L}  = \mathcal{L}_{clf} + \mathcal{L}_{d} $$

\subsubsection{Maximum   Mean   Discrepancy (MMD) Minimization}

Similar to DANN, MMD-minimization offers an alternative way to measure the discrepancy between source and target distributions. Unlike DANN which estimate the $\mathcal{H}-$divergence between distributions, MMD is defined as the squared distance between the kernel embeddings of marginal distributions in the Reproducing kernel Hilbert Space (RKHS). Formally,
$$\text{MMK}_k(\mathbf{L}, \mathbf{UL})=||\mathbb{E}_p[\phi(x_s)] - \mathbb{E}_p[\phi(x_t)]||^2_{\mathcal{H}_k},$$
where $\mathcal{H}_k$ denotes the RKHS with a kernel k, and $\mathbf{L}, \mathbf{UL}$ are labeled source and unlabeled target distributions. 

\begin{figure}
    	\centering
	{\includegraphics[width=\columnwidth]{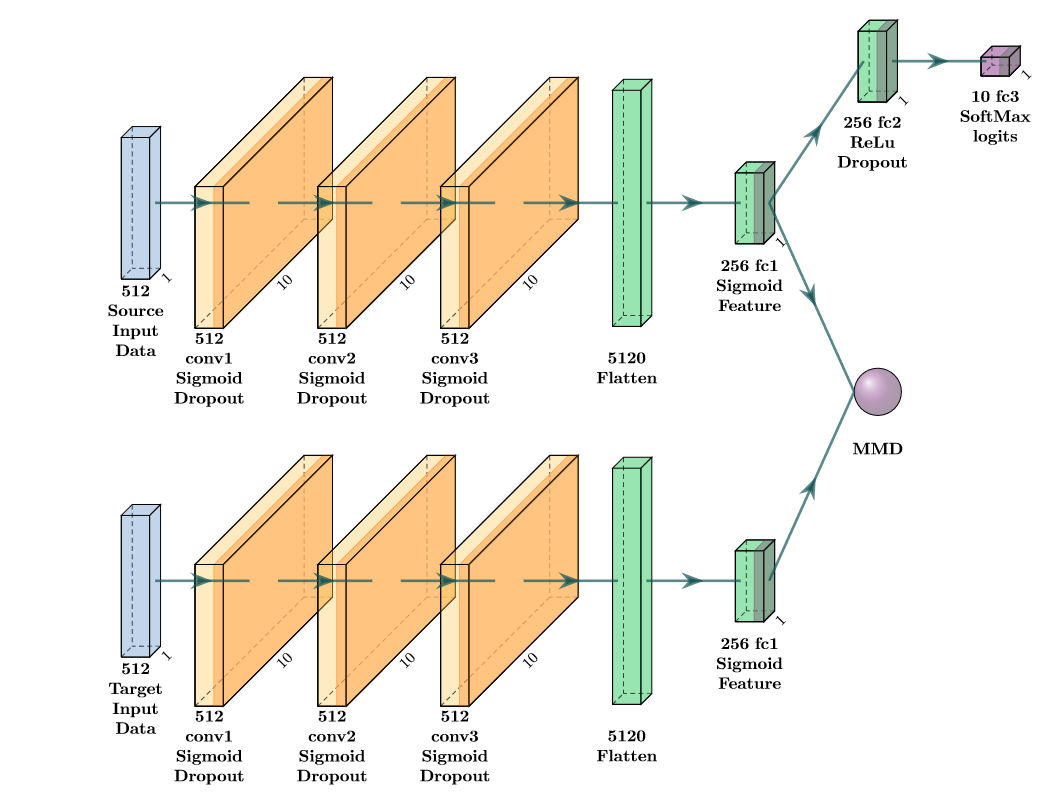}}
	
	\caption{Modified architecture for MMD in comparison experiments.}
	\label{mmd}
\end{figure}
In reality, the choice of kernel used in obtaining these embeddings is crucial to a successful estimation of the discrepancy. Multiple kernels of MMD are usually used to leverage different kernels and provide and effective estimation. 
$$k(x^s, x^t) = \sum_{i=1}^{N_s} k_{\sigma_i}(x_s, x_t),$$
where $ k_{\sigma_i}$ is a Gaussian kernel with width $\sigma_i$. Following the settings in previous MMD works in fault diagnosis~\cite{li2019multi}, we adopt Gaussian kernel widths of 1, 2, 4, 8, and 16. Previous works have shown that this choice of kernels with an equal weight is sufficient enough for our specific task. 

The multi kernel MMD loss is then used as an additional loss along with the label prediction loss to align the feature distribution between source and target machines:
$$\mathcal{L}  = \mathcal{L}_{clf} + \lambda_{MMD}\mathcal{L}_{MMD}(x_s, x_t) $$
\subsubsection{Adaptive Batch Normalization (AdaBN)~\cite{li2016revisiting}}

\begin{figure}
	\centering
	{\includegraphics[width=\columnwidth]{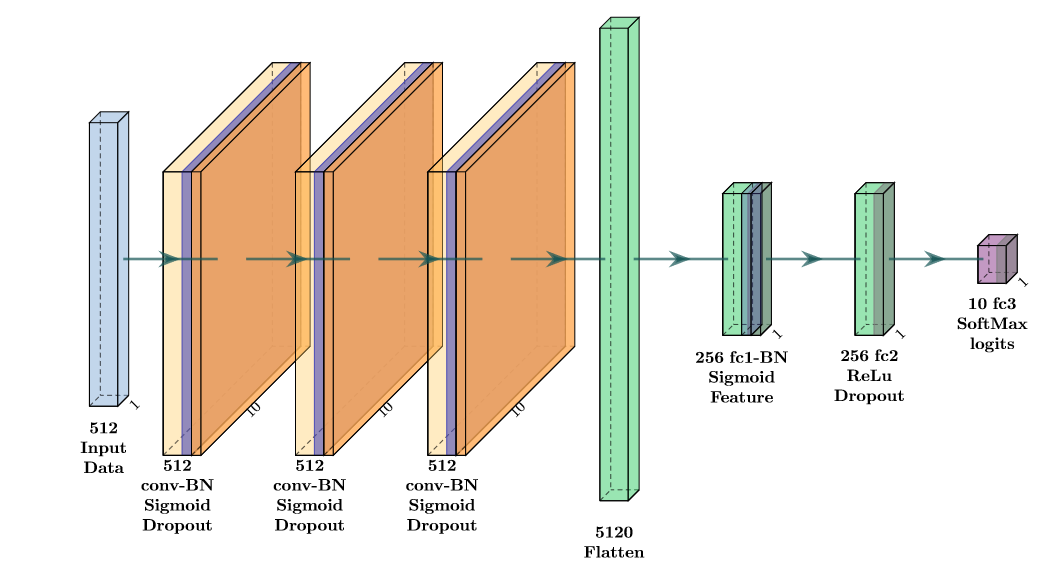}}
	\caption{Modified architecture for AdaBN in comparison experiments.}
	\label{feat-ada}
\end{figure}

Before introducing AdaBN, we briefly review Batch Normalization (BN)\cite{ioffe2015batch}. BN layers are designed to alleviate internal covariate shifting by guaranteeing the input distribution of each layer remains unchanged across different mini-batches. Considering an intermediate representation $x\in\mathbb{R}^{b\times p}$, where $b$ is the batch size and $p$ is the dimension of features. The BN layer transforms a feature by:
$$\hat{x_j} = \frac{x_j - \mathbb{E}[x_{\cdot j}]}{\sqrt{Var[x_{\cdot j}]}}$$
$$y_j = \gamma_j \hat{x_j} + \beta_j, $$
where $j\in \{1...p\}$, and $y$ is the output of the BN layer. $\gamma$ and $\beta$ are parameters to be learned in the training process. The mean and variance statistics are calculated over mini-batch during training, but over the whole population on test time.  

AdaBN is based on the simple assumption that the deterioration of models on the target machine is caused by a distribution discrepancy on intermediate layers. By adding batch normalization layers and replacing BN statistics from source data with those from target data, the distribution difference is expected to be reduced in each layer, thus, increasing the model's performance on the target data. Apart from a small amount of Batch Normalization parameters, AdaBN requires no additional parameters and is easy to implement.

In our AdaBN experiments, the Batch Normalization layers are inserted after each convolutional layer in the feature extractor. After training, we fix $\sigma$, $\beta$, and all other trainable variables, and finetune the batch normalization statistics $\mu$, and $\sigma$ using the target data.

\section{Case Study}
We now present a case study on the CWRU bearing dataset set using the above methods. The case study is designed to make the comparison over methods applicable in realistic fault diagnosis settings. More specifically, we consider the following factors:

\begin{table}[b]
\caption{Class Label Definition for CWRU Dataset }

    \begin{center}
\begin{threeparttable}[b]
        \begin{tabular}{|c|c|c|c|c|c|c|c|c|c|c|}
        \hline
        \textbf{Fault }&\multicolumn{10}{c|}{\textbf{Class Label}} \\
        \cline{2-11} 
        \textbf{} & 0& 1& 2& 3& 4& 5& 6& 7& 8& 9 \\
        \hline
        Loc&NA$^1$& IF& IF&  IF&  BF&BF&BF&OF&OF&OF  \\
        \hline
        Size&0& 7& 14&  21&  7&14&21&7&14&21  \\
        \hline
        \end{tabular}

 \begin{tablenotes}
     \item[1] Fault location not applicable because class 0 is the healthy state.
   \end{tablenotes}
  \end{threeparttable}

    \end{center}
\label{tab1}
\end{table}

\begin{itemize}
	\item The basic backbone architectures are the same across different experiments, so that the capacity of the models does not affect the results.
	\item All experiments share the same pre-processing steps to exclude the effect of the number of samples and augmentation methods.
	\item The different models share a similar budget for hyper-parameter tuning. 
	\item A realistically chosen validation set is used for hyper-parameter tuning. 
\end{itemize}

\subsection{Dataset}
The CWRU bearing dataset~\cite{smith2015rolling} from Bearing Data Center of Case Western Reserve University is used in our experiments. The dataset is chosen because of its availability to the public and its popularity over a large number of previous papers, including studies in domain adaptation. Following the general setup used by most other bearing diagnosis papers, drive end accelerometer data are used in all our experiments. 

Following the label definition setup used by~\cite{li2018cross}, 10 bearing conditions are considered as shown in Table~\ref{tab1}. Three fault types are included: inner race fault (IF), ball fault (BF), and outer race fault (OF). Faults were introduced to the bearings using electro-discharge machining with fault diameters of 7 mils, 14 mils, 21 mils. In total, there are 9 fault states and one healthy state. The dataset was originally collected at 12 and 48 kHz. In all our experiments, we make use of data at 12 kHz sampling rate. If the data are not available at 12 kHz, we down-sample them to ensure a continuous 12 kHz sampling rate over all data points.

The CWRU dataset comprises data from four different loads, which we treat as four different working conditions $\{0, 1, 2, 3\}$. The domain adaptation is applied across different loads. In this section we denote Task $0\xrightarrow{}1$ as the setup where source domain is the working load 0 and target domain is the working load 1.

\begin{table*}[t]
\centering

\caption{Domain Adaptation Results.}

   \begin{threeparttable}[b]
\begin{tabular}{|c|c|c|c|c|c|c|c|c|
a |a |a |}
\hline
Setup    & \multicolumn{2}{c|}{Baseline}        & \multicolumn{2}{c|}{DANN}       & \multicolumn{2}{c|}{MMD}        & \multicolumn{2}{c|}{AdaBN}        & AdaBN~\cite{zhang2017new}$^1$ & MMD-ML~\cite{li2019multi}${^1} {^2} {^3}$& A2CNN~\cite{zhang2018adversarial}${^1} {^2}$ \\ \hline
         & Mean${^4}$     & Max${^4}$     & Mean${^4}$   & Max${^4}$     & Mean${^4}$    & Max${^4}$     & Mean${^4}$   & Max${^4}$     & Reported${^5}$   & Reported${^5}$       & Reported${^5}$     \\ \hline
Task 0-1 & 93.49  & 95.45  & 98.76 & 99.30  & 99.38  & 99.50  & 98.87  & 99.35  & -     & 99.56 & -     \\ \hline
Task 0-2 & 93.65  & 95.15  & 99.96 & 100.00 & 99.98  & 100.00 & 99.30  & 99.75  & -     & 99.48 & -     \\ \hline
Task 0-3 & 91.02  & 94.75  & 99.81 & 100.00 & 100.00 & 100.00 & 99.75  & 99.80  & -     & 99.17 & -     \\ \hline
Task 1-0 & 97.93  & 98.30  & 98.73 & 99.05  & 99.31  & 99.40  & 98.83  & 99.05  & -     & -     & -     \\ \hline
Task 1-2 & 100.00 & 100.00 & 99.96 & 100.00 & 99.98  & 100.00 & 99.95  & 99.95  & 99.40 & -     & 99.99 \\ \hline
Task 1-3 & 98.26  & 99.35  & 99.65 & 99.80  & 99.97  & 100.00 & 99.82  & 99.85  & 93.40 & -     & 99.30 \\ \hline
Task 2-0 & 91.63  & 94.40  & 97.70 & 98.25  & 98.61  & 98.65  & 95.89  & 96.40  & -     & -     & -     \\ \hline
Task 2-1 & 97.09  & 98.05  & 98.40 & 98.45  & 98.52  & 98.60  & 97.83  & 98.15  & 97.50 & -     & 98.18 \\ \hline
Task 2-3 & 99.78  & 100.00 & 99.82 & 99.95  & 100.00 & 100.00 & 100.00 & 100.00 & 97.20 & -     & 99.90 \\ \hline
Task 3-0 & 87.96  & 88.25  & 97.62 & 97.85  & 98.72  & 98.90  & 89.27  & 90.30  & -     & 97.58 & -     \\ \hline
Task 3-1 & 89.42  & 91.15  & 98.41 & 98.50  & 98.53  & 98.60  & 94.42  & 95.10  & 88.30 & 98.61 & 97.93 \\ \hline
Task 3-2 & 99.65  & 99.90  & 99.98 & 100.00 & 100.00 & 100.00 & 99.95  & 99.95  & 99.90 & 99.05 & 99.99 \\ \hline
Average  & 94.99  & 96.23  & 99.07 & 99.26  & 99.42  & 99.47  & 97.82  & 98.14  & -     & -     & -     \\ \hline
Train time  & \multicolumn{2}{c|}{84 s}        & \multicolumn{2}{c|}{177 s}       & \multicolumn{2}{c|}{266 s}        & \multicolumn{2}{c|}{133 s}  & -     & -         & -       \\ \hline
Parameter  & \multicolumn{2}{c|}{1379998}        & \multicolumn{2}{c|}{2694816}       & \multicolumn{2}{c|}{1379998}        & \multicolumn{2}{c|}{1380570}  & -     & -         & -       \\ \hline
\end{tabular}

   \begin{tablenotes}
     \item[1] Grey columns are results reported by relevant papers. Results may not be directly comparable because of different backbone and pre-processing steps. 
     \item[2] ML in MMD-ML stands for multi-layer, where  MMD is applied on multiple intermediate layers.
     \item[3] MMD-ML~\cite{li2019multi} results are estimated from Figure 4 in the original paper. 
     \item[4] Reproduced numbers are based on average or max numbers over five runs. 
     \item[5] \cite{zhang2017new} and \cite{zhang2018adversarial} did not specify the number of runs. MMD-ML~\cite{li2019multi} results are based on average over ten runs. 
   \end{tablenotes}
  \end{threeparttable}
\label{tab2}
\end{table*}

\subsection{Preprocessing}
We mostly follow the same preprocessing steps as~\cite{li2018cross}. It consists in truncating the signal first 120,000 points. They are divided into 200 sequences of 1024 points with some overlap between sequences. Using the Fast Fourier Transform, each sequence is converted into a vector of 512 Fourier coefficients. All data are then normalized by a simple normalization factor. The normalization factor is chosen between $\{1, 8, 64, 512\}$. Normalization factor for all experiments are the same. It is determined by using the one that maximizes the performance on the source-only baseline on the validation task.

\subsection{Baseline}
To fairly evaluate domain adaptation methods for fault diagnosis applications, a strong baseline is critical. In our case study, we use the feature extractor along with a basic classifier as our baseline as shown in Fig~\ref{feat}, and train it using only source data. No additional target data are used in the baseline. To choose the hyper-parameters, we use the task $0\xrightarrow{}3$ as validation task to tune all models, because it is one of the most difficult tasks among all the transfer pairs. We use Adam optimizer with a learning rate of 0.0002. The general hyper-parameters are fixed and shared by all other experiments once the baseline model is optimized according to the validation task.  

\subsection{Budgets for Method-specific Hyper-parameters}
To fairly compare the different methods, equivalent budgets for hyper-parameters should be used for all models. For DANN models, we use $\{0.1, 1, 10\}$ as the pool of hyper-parameters for gradient reverse factor $\lambda_d$. For MMD models, we use $\{0.1, 1, 10\}$, for the MMD discrepancy weight $\lambda_{MMD}$. AdaBN does not require any additional hyper-parameter. We train all models for 2000 Epochs.

\subsection{Experimental Environment}
NVIDIA GTX 1080 is used for all experiments. The main framework is written using Python and Tensorflow. We run all experiments five times and report average and maximum accuracy to reflect the model performance and stability.

\subsection{Experiment Results}

In the following the experimental results of different domain adaptation methods are reported.

\subsubsection{Model Performance}
By carefully tuning the basic backbone using the validation task, we report the average accuracy of 94.99\% for CWRU dataset. We argue that this is a rather strong baseline, as it is stronger than that reported in previous papers~\cite{zhang2018deep,  zhang2018adversarial, li2018cross}, and close to some of the results reported in studies applying domain adaptation~\cite{li2018cross}. By providing a strong baseline, our evaluation  reflects more fairly the effectiveness and applicability of the discussed domain adaptation methods. 

Under the assumption of availability of unlabeled data on target domain, all domain adaptation methods discussed in this paper are able to improve model performance. DANN yields very good results achieving over 99.0\% of average accuracy on the target domain, suggesting a meaningful feature alignment and a successful adaptation. Similarly, the MMD approach improves the model performance on target data and achieves an average accuracy of 99.4\% over all tasks. The only drawback of MMD method may arise in cases when the data size is larger that the training time can quadratically increase. AdaBN, as a simple method without any additional parameters, also improves the model performance, though not as significantly as the other methods. The advantage of AdaBN is that it could be easily combined with other domain adaptation methods without increasing the model complexity. 

On the right side of Table~\ref{tab2}, we show results from previous works using similar approaches for comparison. These results are not directly comparable with other columns because each paper uses its own way to prepare the target test set. MMD-ML~\cite{li2019multi} uses a similar MMD setup as ours, except that they apply the MMD loss not only on the feature layer, but also on other intermediate layers.  A2CNN~\cite{zhang2018adversarial} uses adversarial training for domain adaptation, and shares a similar idea as DANN. The key difference between A2CNN and DANN is that A2CNN implementation does not use a reverse gradient layer, but utilizes a two-step training for classifiers and discriminators. This requires a more careful tuning of the training strategy.  

The missing cells in Table~\ref{tab2} filled with $-$ mean that the original papers do not report results on these tasks. In this paper, we report model performance on all available adaptation tasks on the CWRU dataset. We believe by doing so, it provides a better and fairer comparison among domain adaptation methods for fault diagnosis.

\subsubsection{Model Efficiency}
Model efficiency is crucial in reality for fault diagnosis applications, as computational resources may be limited.  In Table~\ref{tab2}, for each method, we report a training time for 2000 Epochs and model complexity in terms of trainable parameters. We believe that this brings more insight in characteristics of these methods. As explained in the method section, AdaBN is the fastest domain adaptation method among all three we introduced in this paper. A small amount of extra parameter is introduced by batch normalization layers in the AdaBN training, and asks for comparable small amount of additional time. MMD methods, on the other hand, ask for no additional trainable parameter, but require a significantly higher amount of time for training. The additional time results from the time-consuming procedure of MMD estimation in every training iteration. One additional problem of MMD-related methods is its quadratic time complexity with regard to sample size. This limits its application in a more general scenario, where more training points are available. DANN method requires more parameters because of the additional domain classifier. The training time, however, is significantly  smaller than that of the MMD methods. By using gradient reverse layers, the adversarial training procedures fit into the standard gradient descent training of the neural networks, and thus the $\mathcal{H}$-divergence can be estimated efficiently. 

\subsubsection{Discussion}
DANN, MMD, and AdaBN are all able to improve the model performance on the target task. AdaBN requires few additional parameters and provides a moderate adaptation ability with a minimum additional computational cost. MMD, on the other side, yields the best results for the bearing dataset at the largest computational cost. It also has a potential problem on efficiently dealing with larger training sets. The DANN method we introduced to fault diagnosis from \cite{ganin2014unsupervised} can be considered as a good trade-off between accuracy and computational power. It provides us with competitive results with the help of a reasonable amount of additional computational cost.

\section{Conclusion}
In the present paper, we proposed to use DANN, an adversarial domain adaptation method, for supervised fault diagnosis tasks. We compared its performance with two other domain adaptation methods. To enable a fair comparison between the methods and to to evaluate their applicability and effectiveness for fault diagnosis problems in reality, we proposed a unified experimental procedure. All of the methods applied in this case study were able to improve model performance on target data, suggesting these domain adaptation methods provide an added value to fault diagnosis problems in real applications. DANN method provides competitive results using significantly less training time comparing to MMD, and yields superior results over AdaBN.

%\section*{Acknowledgment}

\bibliographystyle{IEEEtran}
\bibliography{conference_041818}

\begin{thebibliography}{10}
\providecommand{\url}[1]{#1}
\csname url@samestyle\endcsname
\providecommand{\newblock}{\relax}
\providecommand{\bibinfo}[2]{#2}
\providecommand{\BIBentrySTDinterwordspacing}{\spaceskip=0pt\relax}
\providecommand{\BIBentryALTinterwordstretchfactor}{4}
\providecommand{\BIBentryALTinterwordspacing}{\spaceskip=\fontdimen2\font plus
\BIBentryALTinterwordstretchfactor\fontdimen3\font minus
  \fontdimen4\font\relax}
\providecommand{\BIBforeignlanguage}[2]{{%
\expandafter\ifx\csname l@#1\endcsname\relax
\typeout{** WARNING: IEEEtran.bst: No hyphenation pattern has been}%
\typeout{** loaded for the language `#1'. Using the pattern for}%
\typeout{** the default language instead.}%
\else
\language=\csname l@#1\endcsname
\fi
#2}}
\providecommand{\BIBdecl}{\relax}
\BIBdecl

\bibitem{zhang2016understanding}
C.~Zhang, S.~Bengio, M.~Hardt, B.~Recht, and O.~Vinyals, ``Understanding deep
  learning requires rethinking generalization,'' \emph{arXiv preprint
  arXiv:1611.03530}, 2016.

\bibitem{pan2011domain}
S.~J. Pan, I.~W. Tsang, J.~T. Kwok, and Q.~Yang, ``Domain adaptation via
  transfer component analysis,'' \emph{IEEE Transactions on Neural Networks},
  vol.~22, no.~2, pp. 199--210, 2011.

\bibitem{long2015learning}
M.~Long, Y.~Cao, J.~Wang, and M.~I. Jordan, ``Learning transferable features
  with deep adaptation networks,'' \emph{arXiv preprint arXiv:1502.02791},
  2015.

\bibitem{li2016revisiting}
Y.~Li, N.~Wang, J.~Shi, J.~Liu, and X.~Hou, ``Revisiting batch normalization
  for practical domain adaptation,'' \emph{arXiv preprint arXiv:1603.04779},
  2016.

\bibitem{ganin2014unsupervised}
Y.~Ganin and V.~Lempitsky, ``Unsupervised domain adaptation by
  backpropagation,'' \emph{arXiv preprint arXiv:1409.7495}, 2014.

\bibitem{saito2018maximum}
K.~Saito, K.~Watanabe, Y.~Ushiku, and T.~Harada, ``Maximum classifier
  discrepancy for unsupervised domain adaptation,'' in \emph{Proceedings of the
  IEEE Conference on Computer Vision and Pattern Recognition}, 2018, pp.
  3723--3732.

\bibitem{zhang2017new}
W.~Zhang, G.~Peng, C.~Li, Y.~Chen, and Z.~Zhang, ``A new deep learning model
  for fault diagnosis with good anti-noise and domain adaptation ability on raw
  vibration signals,'' \emph{Sensors}, vol.~17, no.~2, p. 425, 2017.

\bibitem{zhang2018adversarial}
B.~Zhang, W.~Li, M.~Zhang, and Z.~Tong, ``Adversarial adaptive 1-d
  convolutional neural networks for bearing fault diagnosis under varying
  working condition,'' \emph{arXiv preprint arXiv:1805.00778}, 2018.

\bibitem{li2019multi}
X.~Li, W.~Zhang, Q.~Ding, and J.-Q. Sun, ``Multi-layer domain adaptation method
  for rolling bearing fault diagnosis,'' \emph{Signal Processing}, vol. 157,
  pp. 180--197, 2019.

\bibitem{li2015multimodal}
C.~Li, R.-V. Sanchez, G.~Zurita, M.~Cerrada, D.~Cabrera, and R.~E. V{\'a}squez,
  ``Multimodal deep support vector classification with homologous features and
  its application to gearbox fault diagnosis,'' \emph{Neurocomputing}, vol.
  168, pp. 119--127, 2015.

\bibitem{zhai2016deep}
S.~Zhai, Y.~Cheng, W.~Lu, and Z.~Zhang, ``Deep structured energy based models
  for anomaly detection,'' \emph{arXiv preprint arXiv:1605.07717}, 2016.

\bibitem{jia2016deep}
F.~Jia, Y.~Lei, J.~Lin, X.~Zhou, and N.~Lu, ``Deep neural networks: A promising
  tool for fault characteristic mining and intelligent diagnosis of rotating
  machinery with massive data,'' \emph{Mechanical Systems and Signal
  Processing}, vol.~72, pp. 303--315, 2016.

\bibitem{tamilselvan2013failure}
P.~Tamilselvan and P.~Wang, ``Failure diagnosis using deep belief learning
  based health state classification,'' \emph{Reliability Engineering \& System
  Safety}, vol. 115, pp. 124--135, 2013.

\bibitem{li2018cross}
X.~Li, W.~Zhang, and Q.~Ding, ``Cross-domain fault diagnosis of rolling element
  bearings using deep generative neural networks,'' \emph{IEEE Transactions on
  Industrial Electronics}, 2018.

\bibitem{zhang2018deep}
W.~Zhang, C.~Li, G.~Peng, Y.~Chen, and Z.~Zhang, ``A deep convolutional neural
  network with new training methods for bearing fault diagnosis under noisy
  environment and different working load,'' \emph{Mechanical Systems and Signal
  Processing}, vol. 100, pp. 439--453, 2018.

\bibitem{saenko2010adapting}
K.~Saenko, B.~Kulis, M.~Fritz, and T.~Darrell, ``Adapting visual category
  models to new domains,'' in \emph{European conference on computer
  vision}.\hskip 1em plus 0.5em minus 0.4em\relax Springer, 2010, pp. 213--226.

\bibitem{tzeng2017adversarial}
E.~Tzeng, J.~Hoffman, K.~Saenko, and T.~Darrell, ``Adversarial discriminative
  domain adaptation,'' in \emph{Proceedings of the IEEE Conference on Computer
  Vision and Pattern Recognition}, 2017, pp. 7167--7176.

\bibitem{michau2019}
G.~Michau and O.~Fink, ``Unsupervised {{Fault Detection}} in {{Varying
  Operating Conditions}},'' in \emph{Proceedings of the 2019 {{IEEE
  International Conference}} on {{Prognostics}} and {{Health Management}}},
  2019.

\bibitem{goodfellow2014generative}
I.~Goodfellow, J.~Pouget-Abadie, M.~Mirza, B.~Xu, D.~Warde-Farley, S.~Ozair,
  A.~Courville, and Y.~Bengio, ``Generative adversarial nets,'' in
  \emph{Advances in neural information processing systems}, 2014, pp.
  2672--2680.

\bibitem{ben2010theory}
S.~Ben-David, J.~Blitzer, K.~Crammer, A.~Kulesza, F.~Pereira, and J.~W.
  Vaughan, ``A theory of learning from different domains,'' \emph{Machine
  learning}, vol.~79, no. 1-2, pp. 151--175, 2010.

\bibitem{cortes2011domain}
C.~Cortes and M.~Mohri, ``Domain adaptation in regression,'' in
  \emph{International Conference on Algorithmic Learning Theory}.\hskip 1em
  plus 0.5em minus 0.4em\relax Springer, 2011, pp. 308--323.

\bibitem{ioffe2015batch}
S.~Ioffe and C.~Szegedy, ``Batch normalization: Accelerating deep network
  training by reducing internal covariate shift,'' \emph{arXiv preprint
  arXiv:1502.03167}, 2015.

\bibitem{smith2015rolling}
W.~A. Smith and R.~B. Randall, ``Rolling element bearing diagnostics using the
  case western reserve university data: A benchmark study,'' \emph{Mechanical
  Systems and Signal Processing}, vol.~64, pp. 100--131, 2015.

\end{thebibliography}

%\begin{figure}[htbp]
%\centerline{\includegraphics{fig1.png}}
%\caption{Example of a figure caption.}
%\label{fig}
%\end{figure}
\end{document}